\documentclass{article}

\usepackage{PRIMEarxiv}
\usepackage[T1]{fontenc}    
\usepackage{hyperref}       
\usepackage{url}            
\usepackage{booktabs}       
\usepackage{amsfonts}       
\usepackage{nicefrac}       
\usepackage{amsmath}
\usepackage{microtype}      
\usepackage{lipsum}

\usepackage{float}
\usepackage{algorithm}
\usepackage{algpseudocode}
\usepackage{fancyhdr}       
\usepackage{graphicx}       
\graphicspath{{media/}}     
\usepackage{xcolor}
\usepackage{multirow}
\title{Verbal Werewolf: Engage Users with Verbalized Agentic Werewolf Game Framework
}

\author{
  Qihui Fan, Wenbo Li, Enfu Nan, Yixiao Chen, Lei Lu, Pu Zhao, Yanzhi Wang \\
  Northeastern University \\
  Boston, MA \\
  \texttt{\{fan.qih, li.wenbo6,  nan.e, chen.yixia, lu.lei1, p.zhao, yanz.wang\}@northeastern.edu}
}

\begin{document}
\maketitle

\begin{abstract}
The growing popularity of social deduction games has created an increasing need for intelligent frameworks where humans can collaborate with AI agents, particularly in post-pandemic contexts with heightened psychological and social pressures. Social deduction games like Werewolf, traditionally played through verbal communication, present an ideal application for Large Language Models (LLMs) given their advanced reasoning and conversational capabilities. Prior studies have shown that LLMs can outperform humans in Werewolf games, but their reliance on external modules introduces latency that left their contribution in academic domain only, and omit such game should be user-facing. We propose \textbf{Verbal Werewolf}, a novel LLM-based Werewolf game system that optimizes two parallel pipelines: gameplay powered by state-of-the-art LLMs and a fine-tuned Text-to-Speech (TTS) module that brings text output to life. Our system operates in near real-time without external decision-making modules, leveraging the enhanced reasoning capabilities of modern LLMs like \texttt{DeepSeek V3} to create a more engaging and anthropomorphic gaming experience that significantly improves user engagement compared to existing text-only frameworks.
\end{abstract}


\section{Introduction}
There is an increasing need for intelligent social deduction game frameworks which one or few people can play with collaboration with AI due to the post-pandemic psychological pressure and heavier social burden. Social deduction games, Werewolf as one of the most typical ones, are often played in forms of verbal communication, which we argue the large language models(LLMs) should be excelled at\cite{huang2022language}. Previous works have already confirmed that LLMs have certain probability to defeat human players in the werewolf games \cite{shibata2023playingwerewolfgameartificial} with additional model for estimating the probability of winning the game. Other further works similarly have also been shown that high-performance LLMs with external modules for strategy optimizations can play the werewolf games strategically without fine-tuning the language models \cite{xu2023exploring, wu2024enhancereasoninglargelanguage}. However, these related works are heavily dependent on external modules for decision-makings and often demonstrate the game in simple language only. In addition, external modules often bring excessive latency, and disable the anthropopathic type-writing flush response that only constrains their framework at an academic level, overlooked the low-delay and anthropopathic nature as an engaging social deduction game. 

In light of this, we propose \textbf{Verbal Werewolf} to further explore a more engaging werewolf game framework by optimizing two pipelines, one for the gameplay powered by the most state-of-the-art LLMs, and the other one for our fine-tuned Text to Speech(TTS) module that brings plain text output to live. We demonstrate the Verbal Werewolf progresses both pipeline in parallel and in nearly real-time, improving the user experience drastically.

\section{Implementation}
We demonstrate the Verbal Werewolf framework in mainly two pipelines: Game Play that manages the game flow, such roles assignments, LLM queries, and TTS Module that transforms sentence-by-sentence text output to designated audio outputs and manage the aduio play flow.
\begin{figure}[h]
  \centering
  \includegraphics[width=1\linewidth]{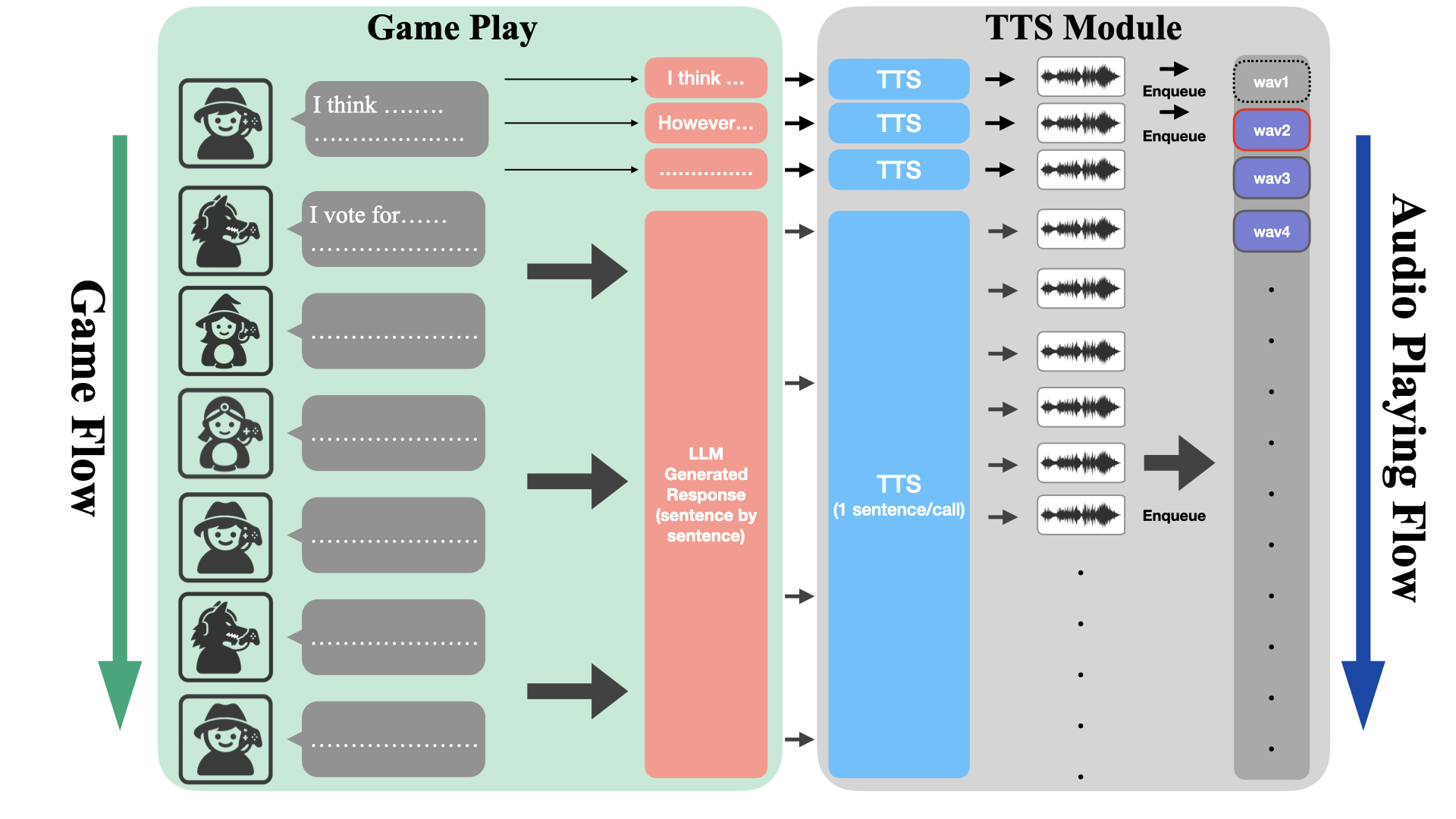}  
  \caption{An overview of parallel processing design of LLM and TTS audio generations.}
  \label{fig:your-label}
\end{figure}
\subsection{Game Play}
\textbf{Game Flow} We present the werewolf game in the simplest setting\cite{xu2023language} involving only \textit{werewolf}, \textit{villager}, \textit{seer}, and \textit{witch} roles. The Game Play mainly manages the whole game flow by initially assign different roles randomly, and we mainly divide one round of play into \texttt{daytime}, when they should \texttt{reason} then \texttt{vote} who are more suspicious being werewolves, and \texttt{nighttime}, when different players executes their assigned actions per their roles in the sequence of \texttt{\{werewolves-seer-witch\}} We formally define their action space as follows:
\begin{table}[h]
\centering
{\small
\begin{tabular}{|l|l|l|l|}
\hline
\textbf{Role} & \textbf{\texttt{Daytime} Actions} & \textbf{\texttt{Nighttime} Actions} & \textbf{Props} \\
\hline
Villager & \multirow{4}{*}{\texttt{reason}, \texttt{vote}} & & \\
\cline{1-1}\cline{3-4}
Werewolf & & \texttt{kill} & \\
\cline{1-1}\cline{3-4}
Seer & & \texttt{reveal} & \\
\cline{1-1}\cline{3-4}
Witch & & \texttt{night} & \texttt{cures}, \texttt{poisons} \\
\hline
\end{tabular}
}
\end{table}

\noindent In more detail, werewolves each round can \texttt{kill} a specific active player when this action is called, with the system selecting a player from their available choices if they differ. The Seer is able to \texttt{reveal} the assigned role of a particular player each round. The Witch can additionally \texttt{cure} the player that werewolves attempt to eliminate each round if the witch still possesses \texttt{cures}, or \texttt{poison} a particular player of their choice if the witch still has \texttt{poisons} available --- all within one concatenated \texttt{night} action.

\noindent\textbf{AI Players} LLM-powered agent systems excel at various aspects, from dynamically following and executing instructions to interacting with environments that require solid reasoning capabilities. Their performance relies on the capabilities of the integrated LLMs. We assume that nowadays LLMs are capable of playing the Werewolf or a similar form of game without additional complex design such as Chain-of-Thoughts (CoT) and ReAct prompting, or external experience pool\cite{wang2023avalon}.

\noindent We integrated DeepSeek V3, one of the most state-of-the-art LLMs, as it offers a practical balance that provides strong reasoning capabilities, one is distilled from the reasoning-focused DeepSeek R1, while avoiding the longer response times of DeepSeek R1 and the higher costs of OpenAI models. We have also integrated portals for other LLMs available from HuggingFace, llama.cpp, and OpenAI to facilitate future experiments and development.

\noindent Our approach is straightforward: we prompt the LLM by providing information about the players to be role-played, a description of the action space, guidance on reply format, and minimal guidance about the game rules.

\subsection{Text-to-Speech(TTS) Module}
We fine-tuned \texttt{GPT-SoVITS} \cite{gpt-sovits}, a state-of-the-art TTS model that already excels at fast one-shot voice cloning from brief audio samples (5-10 seconds) with universal weights, by creating specialized model weights for each of 8 celebrities. While the original model provides excellent general-purpose voice synthesis, our fine-tuned versions achieve superior fidelity in reproducing each celebrity's unique vocal characteristics, bringing their familiar voices into the game with enhanced quality and improving the overall player experience.

\noindent Combining Game Flow and TTS modules, our agentic system employs parallel processing threads to achieve real-time voice interaction. As the Game Play generates AI player responses token by token, the TTS thread concurrently processes every half sentence, synthesizing audio and queuing playback. This streaming architecture eliminates traditional wait-for-completion bottlenecks brought by both LLMs and TTS models, allowing AI players to begin speaking while still formulating their complete responses, creating natural conversational flow that mirrors human gameplay dynamics.

\section{Qualitative Analysis}
\paragraph{Overall Latency} During qualitative testing, we observed minor latency only at the beginning of the discussion phase, typically in the first few sentences spoken by the first player to respond. This slight delay stems from the initial overhead of triggering TTS generation and starting audio playback. However, once initialized, the system runs smoothly, as the audio playback duration generally exceeds the time required for both LLM generation and TTS synthesis. This allows subsequent outputs to be processed in parallel and queued efficiently, resulting in consistently low-latency, natural interactions for the remainder of the game flow.

\paragraph{AI Players' Strategies and Performance} Across qualitative evaluations, our Werewolf game framework reveals that advanced LLM role-play exhibit strong autonomous reasoning and strategic gameplay, even without reliance on external tools. These agents demonstrate role-consistent behaviors, such as cautious information disclosure by Seers, resource-aware decision-making by Witches, and persuasive deception by Werewolves, indicating an emergent understanding of social deduction dynamics.

Notably, these models often take initiative in guiding discussions, shift suspicion based on evolving context, and modulate their language style to influence or evade others. Such behaviors suggest not only role awareness but also a degree of leadership and social alignment typically associated with experienced human players.

In contrast, when running less capable LLMs (e.g., smaller-scale or quantized versions), we observe a significant drop in coherence and strategic consistency. These models tend to produce repetitive, generic statements, fail to adapt to game state changes, and often overlook their special abilities entirely. Their discussions lack depth, and they rarely sustain deception or form persuasive arguments.

These findings reinforce that high-capacity LLMs can serve as autonomous, strategic agents in multi-role, socially interactive environments, demonstrating emergent leadership, coordination, and deception purely through language modeling.

\paragraph{API Caching} During our qualitative benchmarks, we observed that interrupting the game flow, specifically during the first night phase and immediately starting a new game could result in the API returning text generated for the previous session. This caused inconsistencies with the current game’s context. Initially, we suspected the cause is from improper handling of game logs or prompt construction. However, targeted testing with smaller local LLMs (e.g., \texttt{QWen 1.5B Quant 8}) failed to reproduce the issue, suggesting the problem was not on the framework side. We ultimately attributed this behavior to the API server, which, although not documented, appears to compare newly queried prompts with prior ones and return cached generations if the similarity is deemed sufficiently high. 
\paragraph{AI Hallucination} We also observed hallucinations in the more recent version of \texttt{DeepSeek-V3-0324}, where the model role-played game characters by associating them with real-world public figures and made assumptions based on those identities. For instance, the model might consider Jack Ma suspicious solely because he is a prominent businessman, or deem Donald Trump untrustworthy due to his political background, while portraying Jay Chou as less suspicious because of his reputation in the music industry. This behavior persisted even after we carefully tuned the prompts to mitigate such biases. However, the issue occurred less frequently when using an earlier version of the model (\texttt{DeepSeek-V3-1224}) and other LLMs such as \texttt{ChatGPT 4o} even when without the specific prompt tuning.

\section{Limitations}
While our system demonstrates promising autonomous behavior in LLM-driven Werewolf gameplay, several limitations remain. First, it is not a fully agentic architecture, since our design does not incorporate external tools such as retrieval-augmented generation (RAG), persistent memory, or structured knowledge bases. All reasoning is conducted via in-context learning, which makes the system inherently constrained by the context length limitations of different LLMs. As the game progresses, earlier dialogue and decisions may be truncated, potentially impacting consistency, recall, and long-term strategic reasoning.

Second, the current implementation supports only a subset of standard Werewolf roles (e.g., Seer, Witch, Werewolf, Villager). Expanding the system to include more complex roles (e.g., Hunter, Cupid, Guard) would introduce deeper inter-dependencies and increase the difficulty of role alignment, offering a stronger test of agent adaptability.

Third, if our quantitative evaluations primarily rely on win/loss outcomes, such metric can introduce bias. A team victory may reflect the opposing side’s failure rather than superior strategic behavior. Future evaluations should incorporate finer-grained behavioral metrics such as reasoning depth, influence in discussion, and success in deception or alignment as a valid and comprehensive quantitative evaluations.

\section{Conclusion}
Our framework demonstrates that large language models with strong reasoning capabilities can autonomously and effectively participate in the Werewolf game, exhibiting strategic thinking, role-based adaptation, and socially coherent dialogue without the need for external modules or structured planning systems. By integrating a fine-tuned text-to-speech (TTS) module and leveraging a parallel processing design, the system delivers low-latency, natural audio output that enhances immersion and real-time interaction. These components together create a compelling and engaging game experience, highlighting the potential of modern LLMs not only as conversational agents but as active participants in complex, multi-agent social environments.


\clearpage
\bibliographystyle{unsrt}  
\bibliography{references}


\clearpage
\section*{Supplementary Materials}
\addcontentsline{toc}{section}{Supplementary Materials}
This section provides additional implementation details of the Verbal Werewolf, the variables provided to large language model (LLM) agents for role-play, and the prompt design used in different game phases.

\subsection*{Game Flow Implementation}
Below algorithm describes the procedural logic of the Werewolf game, including initialization, role-specific actions during the night phase, and decision-making during the day phase. This step-by-step representation ensures that all players' actions, role constraints, and phase transitions are systematically handled.

\floatname{algorithm}{}
\begin{algorithm}[H]
\renewcommand{\thealgorithm}{} 
\caption{\textbf{Werewolf Game Flow}}
\label{alg:gameflow}
\begin{algorithmic}[1]
\State \textbf{Class} \texttt{Game}
    \State \quad \textbf{Method} \texttt{init(self, num\_AI, num\_Human)}
        \State \quad \quad Set $self.\text{game\_status} \gets \text{True}$
        \State \quad \quad Initialize $self.\text{players} \gets \texttt{self.initialize\_players()}$
        \State \quad \quad Initialize $self.\text{seer\_dict} \gets \{\}$ \Comment{Seer revealed roles}
        \State \quad \quad Set $self.\text{num\_cure} \gets 1$, $self.\text{num\_poison} \gets 1$ \Comment{Default number of cures/poisons}
        \State \quad \quad Initialize logs: $self.\text{log} \gets \{\}$, $self.\text{werewolf\_log} \gets \{\}$
        \State \quad \quad \dots \Comment{Other game setup steps}
    \State \quad \textbf{Method} \texttt{judge(self)} \Comment{Check win/lose conditions}

\State Initialize $game \gets \texttt{Game}(num\_AI, num\_Human)$
\State Initialize $round \gets 1$
\While{$game.\text{game\_status} = \texttt{True}$}
    \State \textit{\# \textbf{Night Phase:} Werewolves, Seer, and Witch take actions}
    \For{each werewolf in $\texttt{game.get\_role('Werewolf')}$}
        \If{werewolf.status = 'Active'}
            \State Call $\texttt{werewolf.kill}(game)$
        \EndIf
    \EndFor
    \If{seer.status = 'Active'}
        \State Call $\texttt{seer.reveal}(game)$
    \EndIf
    \If{witch.status = 'Active'}
        \State Call $\texttt{witch.night}(game)$
    \EndIf
    \State Execute night actions: $\texttt{game.execute}(werewolf\_target, witch\_action)$

    \State \textit{\# \textbf{Day Phase:} Judge, Discussion, and Voting}
    \State Call $game.\texttt{judge()}$
    \If{$game.\text{status} = \texttt{False}$}
        \Return
    \EndIf
    \For{each player in $\texttt{game.get\_active\_players()}$}
        \State Call $\texttt{player.reason}(game)$
    \EndFor
    \For{each player in $\texttt{game.get\_active\_players()}$}
        \State Call $\texttt{player.vote}(game)$
    \EndFor
    \State Call $game.\texttt{process\_voting(votes)}$
    \State Call $game.\texttt{judge()}$
    \If{$game.\text{status} = \texttt{False}$}
        \Return
    \EndIf
    \State $round \gets round + 1$ \Comment{Proceed to next round}
\EndWhile
\end{algorithmic}
\end{algorithm}

\subsection*{Variables for LLM Role-play}
Table~\ref{tab:variables} summarizes the key in-game variables accessible to LLM agents. Access to each variable is restricted based on the player’s role that enables role-specific reasoning and alignment with real-world game constraints.

\begin{table}[H]
\centering
\small
\caption{Key in-game variables accessible to LLM agents, with role-specific visibility constraints.}
\label{tab:variables}
\begin{tabular}{|l|p{10cm}|l|}
\hline
\textbf{Variable} & \textbf{Description} & \textbf{Visibility} \\ \hline
\texttt{active\_players} & List of all currently active players in the game. & All roles \\ \hline
\texttt{werewolves} & Identities of all players assigned the Werewolf role. & Werewolf only \\ \hline
\texttt{seer\_dict} & Seer’s private revelations of other players’ roles. & Seer only \\ \hline
\texttt{werewolf\_log} & History of Werewolf kill actions from previous nights. & Werewolf only \\ \hline
\texttt{witch\_log} & History of the Witch’s actions during the night phase. & Witch only \\ \hline
\texttt{log} & Public game history including eliminations, votes, and discussions. & All roles \\ \hline
\end{tabular}
\end{table}

\subsection*{Prompt Design for LLM Agents}
Below illustrates an example prompt used in the discussion phase. This minimalistic design avoids complex prompting strategies such as CoT or ReAct, relying on the base reasoning capabilities of \texttt{DeepSeek V3}~\cite{deepseekai2025deepseekv3technicalreport}. The prompt provides role-specific information, available actions, and conversational history.

\begin{algorithm}
\renewcommand{\thealgorithm}{} 
\caption{\textbf{Example Prompt for Discussion Phase}}
\label{alg:prompt}
\begin{algorithmic}[1]
\State \texttt{prompt = \Big[}
\State \quad \{"role": "system", "content": "You are a **Werewolf** in a Werewolf Game."\},
\State \quad \{"role": "user", "content": (\}
\State \quad \quad f"You are **\{self.name\}**. Your role is to hide your identity as the Werewolf while reasoning about events and dialogues. "
\State \quad \quad f"You know the following players are also Werewolves: \{', '.join(werewolves)\}. Work with them without revealing identities. "
\State \quad \quad f"You are in round \{round\}. "
\State \quad \quad + conversation\_info
\State \quad \quad f"Previous Werewolf kills: \{werewolf\_log\}. "
\State \quad \quad "Provide concise reasoning to mislead others while ensuring the Werewolves win, without revealing your identity."
\State \quad )
\State \Big]
\end{algorithmic}
\end{algorithm}

\end{document}